# Biomass phenotyping of oilseed rape through UAV multi-view oblique imaging with 3DGS and SAM model

**Authors**

Yutao Shen[1,2], Hongyu Zhou[3], Xin Yang[1,2], Xuqi Lu[1,2], Ziyue Guo[1,2], Lixi Jiang[3], Yong He[1,2], Haiyan Cen[1,2*]

**Affiliations**

[1] College of Biosystems Engineering and Food Science, Zhejiang University, Hangzhou 310058, P.R. China

[2] Key Laboratory of Spectroscopy Sensing, Ministry of Agriculture and Rural Affairs, Hangzhou 310058, P.R. China

[3] College of Agriculture and Biotechnology, Zhejiang University, Hangzhou 310058, P.R. China

* Correspondence: hycen@zju.edu.cn; Tel: +86-571-8898-2527.

**Abstract**

Biomass estimation of oilseed rape is crucial for optimizing crop productivity and breeding strategies. While UAV-based imaging has advanced high-throughput phenotyping, current methods often rely on orthophoto images, which struggle with overlapping leaves and incomplete structural information in complex field environments. This study integrates 3D Gaussian Splatting (3DGS) with the Segment Anything Model (SAM) for precise 3D reconstruction and biomass estimation of oilseed rape. UAV multi-view oblique images from 36 angles were used to perform 3D reconstruction, with the SAM module enhancing point cloud segmentation. The segmented point clouds were then converted into point cloud volumes, which were fitted to ground-measured biomass using linear regression. The results showed that 3DGS (7k and 30k iterations) provided high accuracy, with peak signal-to-noise ratios (PSNR) of 27.43 and 29.53 and training times of 7 and 49 minutes, respectively. This performance exceeded that of structure from motion (SfM) and mipmap Neural Radiance Fields (Mip-NeRF), demonstrating superior efficiency. The SAM module achieved high segmentation accuracy, with a mean intersection over union (mIoU) of 0.961 and an F1-score of 0.980. Additionally, a comparison of biomass extraction models found the point cloud volume model to be the most accurate, with an determination coefficient ($R^2$) of 0.976, root mean square error (RMSE) of 2.92 g/plant, and mean absolute percentage error (MAPE) of 6.81%, outperforming both the plot crop volume and individual crop volume models. This study highlights the potential of combining 3DGS with multi-view UAV imaging for improved biomass phenotyping.

**MAIN TEXT**

**1. Introduction**

Oilseed rape yield is directly linked to the production of edible oil and protein, making it a crucial oil crop in global agriculture and the food industry (Kamran et al., 2020). Early growth and biomass formation are key traits for plant productivity and yield (Basunanda et al., 2010). In winter and semi-winter oilseed rape, early biomass and biomass heterosis are strongly correlated with seed yield (W. Zhao et al., 2016). Oilseed rape plants with higher biomass during the seedling stage typically exhibit stronger growth and nutrient absorption capabilities, effectively suppressing weed growth and leading to greater yield accumulation (Shaw et al., 2021; Shen et al., 2022). Additionally, the green vegetative tissues of oilseed rape during the seedling stage can be consumed as vegetables or used as feed (Yong et al., 2015). Forage oilseed rape offers several advantages, such as rapid growth, high potential yield, low economic cost, and efficient nitrogen and water use (Bernhardsson et al., 2024; Du et al., 2022). Compared to traditional forage crops, oilseed rape has significant benefits, particularly as it grows during the fallow period in autumn and winter, without affecting the production cycle of staple crops, making it a vital component of complementary forage-crop rotation systems (Bennett et al., 2011). Therefore, dissecting the genetic basis of biomass is crucial for improving oilseed rape yield strategies and addressing feed shortages (Jeromela et al., 2017).

Recent advancements in high-throughput phenotyping (HTP) technologies have made it possible to non-invasively and efficiently quantify complex traits of plants at a population scale over time (D. Chen et al., 2014; Langstroff et al., 2022; Watt et al., 2020; Z. Zhao et al., 2023). Coupled with expanding genome sequencing data, phenomics has facilitated the effective identification of genetic factors influencing oilseed rape biomass (Mohd Saad et al., 2021). In a study of winter spelt in Germany, LiDAR-based parameters such as gap fraction and crop height showed strong correlations with aboveground biomass (AGB), with the Pearson correlation coefficient (r) values of -0.82 and 0.77 for wet AGB and -0.70 and 0.66 for dry AGB, respectively (Montzka et al., 2023). By combining vegetation indices (VIs) extracted from multispectral images with a random forest (RF) model, the researchers achieved accurate AGB estimation, with a coefficient of determination ($R^2$), root mean square error (RMSE), and relative root mean square error (rRMSE) of 0.90, 0.21 kg/m², and 14.05%, respectively (Cen et al., 2019). Compared to other optical sensors, RGB cameras not only offer higher spatial resolution but also come at a relatively lower cost, making them suitable for capturing vegetation texture and three-dimensional information, which can be used for biomass estimation (Watawana & Isaksson, 2024). A study on wheat AGB estimation demonstrated that combining UAV-based RGB imagery with elevation data significantly enhanced accuracy. The most precise model, partial least squares regression (PLSR) using plant height (PH) and the color indices (CIS), achieved $R^2$ of 0.81, RMSE of 1248.48 kg/ha, and rRMSE of 21.77% (Wang et al., 2022). In current studies, although RGB cameras are widely used for biomass estimation, most research relies solely on orthophoto RGB images rather than multi-view oblique images. In complex field environments, orthophoto images often encounter issues with overlapping leaves, resulting in incomplete structural information and thus lower prediction accuracy (Costa et al., 2020). Fortunately, with the development of stable UAV flight technology, researchers are now able to capture multi-view oblique RGB images with high spatial resolution. Xiao et al., (2024) innovatively proposed a multi-view UAV image acquisition method called the Cross-circling oblique (CCO) route, which enabled a more comprehensive 3D reconstruction of cotton bolls. In terms of cotton boll counting ($R^2 = 0.92$ vs $R^2 = 0.73$) and yield estimation (seed cotton: $R^2 = 0.7$ vs $R^2 = 0.62$, lint cotton: $R^2 = 0.75$ vs $R^2 = 0.55$), the precision of the CCO-derived point cloud significantly surpassed that of the orthophoto-derived point cloud. Incorporating such multi-view image acquisition techniques into biomass estimation holds the potential to significantly improve the accuracy of structural information capture, ultimately enhancing the precision of yield predictions and other key agronomic traits.

Additionally, the advent of Neural Radiance Fields (NeRF) and 3D Gaussian Splatting (3DGS) allows us to achieve high-quality 3D scene reconstruction from multi-view oblique RGB images, enabling the extraction of more detailed vegetation structure information and enhancing the accuracy of structural parameters such as plant height and canopy density (Y. Chen & Lee, 2024; Huang & Yu, 2024). Due to its ability to achieve high-precision reconstruction, handle complex lighting conditions, and operate without relying on precise geometric information, NeRF is suitable for the 3D reconstruction of most plants. For example, Yang et al. developed a novel method, PanicleNeRF, for high-precision and low-cost 3D reconstruction of rice panicles in the field, achieving strong correlations between panicle volume and both grain number ($R^2 = 0.85$ for indica, 0.82 for japonica) and grain mass, offering a promising solution for high-throughput phenotyping in rice breeding (Yang et al., 2024). At the same time, researchers conducted a comparison of 3D reconstruction accuracy for greenhouse pepper plants using both 3D scanners and NeRF. The comparison revealed that NeRF achieved competitive accuracy, with an average distance error of 0.865 mm between the 3D scanning method and the NeRF method. This indicates that the learning-based NeRF method provided accuracy comparable to 3D scanning while offering improved scalability and robustness (J. Zhao et al., 2024). However, using NeRF for 3D reconstruction typically demands extensive training time to optimize neural network parameters, which incurs substantial computational costs, particularly for high-resolution and large-scale scenes. Although lightweight models like instant-NGP and large-scale approaches such as Block-NeRF have been developed, limitations in reconstruction speed still pose challenges for applying NeRF to large-scale scenarios (Z. Zhang et al., 2024). In response to these challenges, 3DGS has emerged as an alternative that does not rely on volume rendering or complex neural network training, allowing for much faster rendering of 3D structures. In a recent study leveraging instant 3DGS, researchers demonstrated that depth-sensing data, including RGB images, camera poses, and point clouds, can be effectively applied to the 3D reconstruction of a fine-structured indoor plant, providing a streamlined method for point cloud extraction and surface reconstruction (Jäger et al., 2024). Additionally, researchers proposed a pipeline combining super-resolution and 3DGS to enhance the resolution of soybean root structures, achieving high-fidelity 3D reconstructions for the extraction of detailed phenotypic traits (Sun et al., 2024). However, 3DGS has mostly been applied to plant reconstruction in greenhouse settings, and there is currently no dedicated algorithm for directly segmenting 3D Gaussian point clouds. Utilizing 3DGS for field-based 3D reconstruction would significantly enhance the quality of plant reconstruction in open fields and improve the accuracy of extracted phenotypic traits.

Therefore, this study aims to explore the potential of UAV multi-view oblique imaging and advanced 3D reconstruction algorithms for high-throughput biomass phenotyping of oilseed rape. The specific objectives are: (1) to propose a method for acquiring field UAV images using multi-view oblique imaging and validate the performance of 3DGS for field 3D reconstruction, (2) to integrate 3DGS with the Segment Anything Model (SAM) to achieve accurate 3D point cloud segmentation of oilseed rape, and (3) to estimate oilseed rape biomass based on the segmented point clouds and demonstrate improved accuracy compared to the traditional crop volume method based on crop surface models.

## 2. Materials and Methods
### 2.1 Experiment site

The field experiment was conducted during the 2023-2024 growing season at the Jiaxing Academy of Agricultural Sciences, Jiaxing City, Zhejiang Province, China (latitude 29°51′34″ N, longitude 120°41′39″E). The experimental field consisted of 900 plots with each plot area of 1.5 × 1.5 $m^2$ and the spacing of 30 cm between adjacent plots as shown in Fig 1. The experiment was designed with three replications, each containing 300 plots. A total of 296 different breeding materials were transplanted into the field, with 16 oilseed rape seedlings planted per plot. Due to an insufficient number of seedlings, four of the breeding materials were excluded from the experiment. The seedlings were planted on October 1, 2023, and transplanted into the experimental field on October 28, 2023. After transplantation, nitrogen, phosphorus, and potassium fertilizers were applied according to the standard agricultural practices. Routine crop management, including irrigation, weeding, and pest control, was carried out as needed based on weather conditions and field observations to ensure optimal growth.

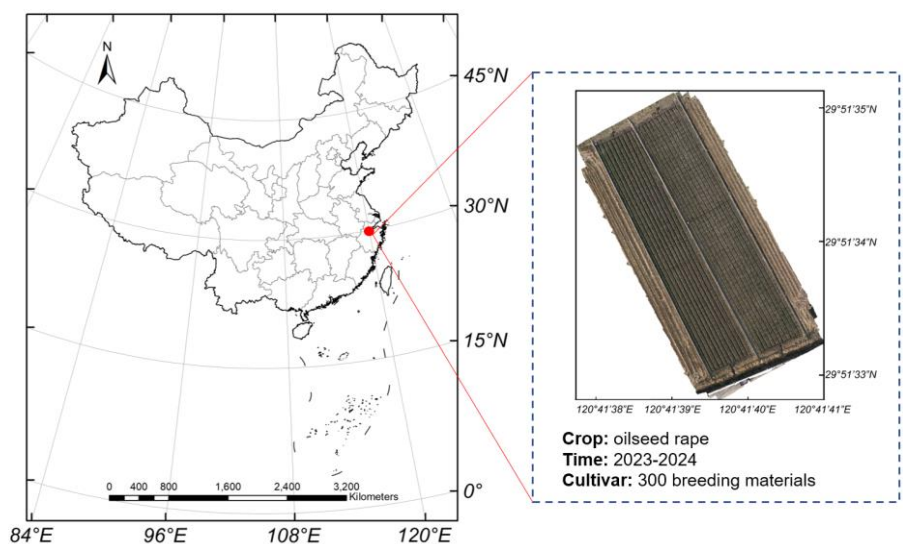

**Fig 1.** Unmanned aerial vehicle (UAV) remote sensing campaign with DJI M300 RTK and PhaseOne

in field experimental sites located at Jiaxing Academy of Agricultural Sciences, Jiaxing City, Zhejiang Province in China in 2023-2024.

## 2.2 UAV-based data collection

2.2.1 UAV data collection timeline

UAV data collection was conducted in three stages during the field experiment. The first dataset, focusing on bare soil imagery, was captured on October 28, 2023. This data was primarily used to extract the digital elevation model (DEM) of the experimental field. The second data collection took place on December 21, 2023, during the seedling stage of oilseed rape, to facilitate high-throughput biomass phenotyping. Starting on March 12, 2024, UAV flights were conducted every 1-2 days to assess the effects of wind speed and crop height on the quality of 3D reconstruction. In total, 33 datasets were collected over the course of the experiment. To ensure accurate georeferencing of the UAV imagery captured at different growth stages, nine ground control points (GCPs) were evenly distributed throughout the field.

2.2.2 Orthophoto and multi-view oblique images acquisition

In this study, a quadcopter UAV (DJI Matrice M300 RTK, Shenzhen, China) equipped with an RGB camera (iXM-100, Phase One, Copenhagen, Denmark) was used for image acquisition. The camera has a CMOS sensor size of 44 × 33 mm and an image resolution of 11664 × 8750 pixels. It is fitted with an 80 mm AF lens (RSM, Phase One, Copenhagen, Denmark, equivalent focal length of approximately 63 mm), offering a field of view (FOV) of about 31.8° horizontally and 21.6° vertically. For the orthophoto images acquisition, the UAV campaigns were conducted at the flight speed of 1.5 m/s with the altitude of 15 m. Both the forward and side overlap rates were set at 80%, resulting in a ground sampling distance (GSD) of approximately 0.65 mm/pixel. Data was stored in IIQ format, and each flight lasted approximately 15 minutes, capturing an estimated 884 nadir images per flight. For the multi-view oblique images acquisition, the UAV's speed was set to 1 m/s, with the altitude of 15 m. The camera pitch angle was fixed at -30°. The 300 plots were divided into 75 groups, each consisting of 4 adjacent plots. For each group, the UAV captured images at 10° intervals around the center point of the 4 plots, resulting in 36 images per group, as illustrated in Fig 2(a). Each flight lasted approximately 75 minutes, capturing a total of 36 × 75 images, all stored in IIQ format. A sample set of 9 images from the 36-image sequence is shown in Fig 3.

UAV flights were typically conducted between 10:00 AM and 2:00 PM local time under clear, cloudless conditions. For the first and second datasets, flights were carried out during periods of minimal wind. For the third dataset, wind conditions were not specifically avoided, but the wind speed was recorded before each flight. During image acquisition, the camera aperture was set to F6.3, with a fixed shutter speed of 1/2000 s. The ISO was set to automatic to ensure proper exposure in varying lighting conditions.

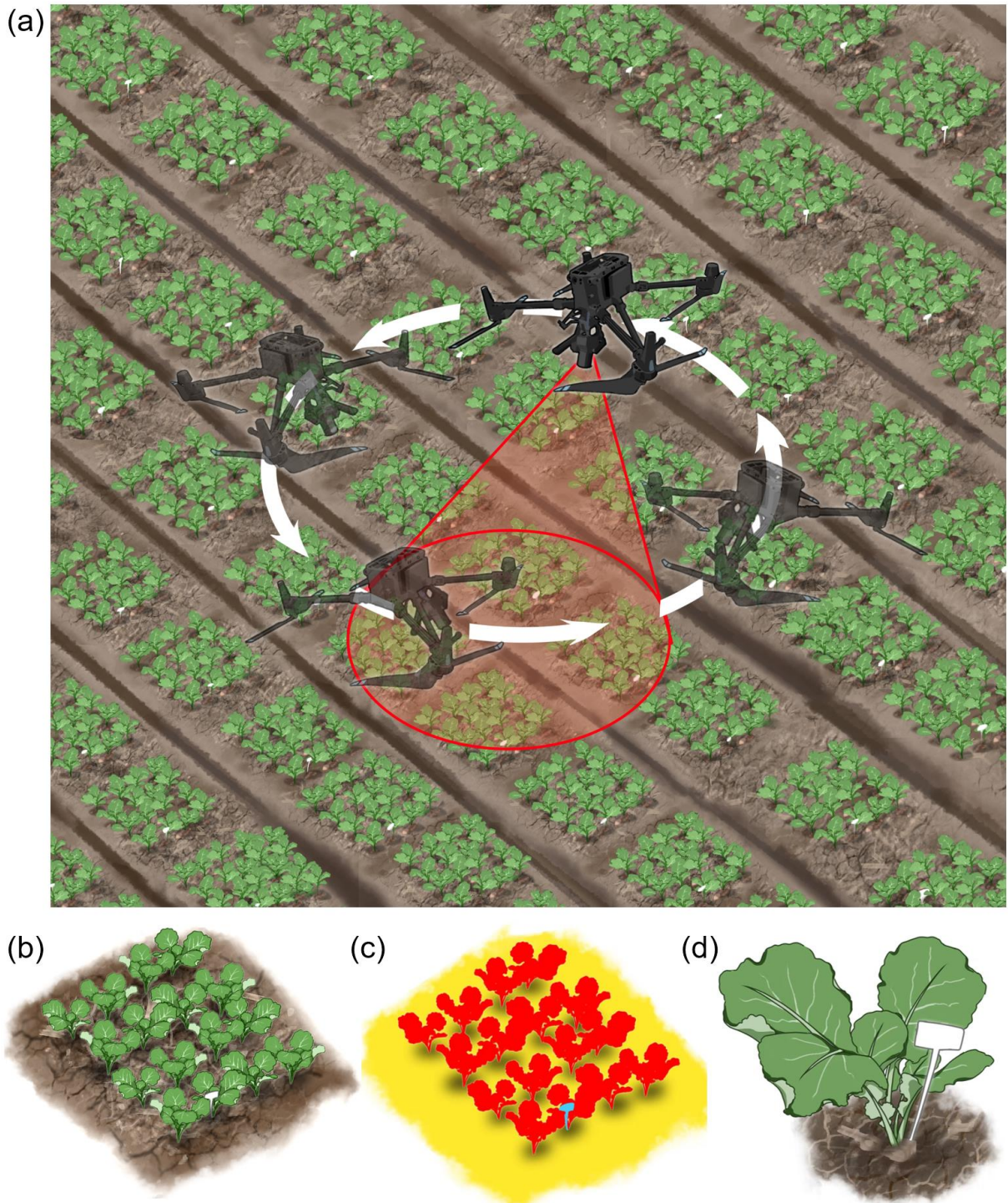

**Fig 2.** Flowchart of biomass estimation using UAV multi-view oblique imaging, 3D gaussian splatting (3DGS), and segment anything model (SAM). (a) Illustration of UAV capturing multi-view oblique images (b) Illustration of 3D field reconstruction (c) Illustration of 3D point cloud segmentation results in the field (d) Illustration of 3D model of a single rapeseed plant.

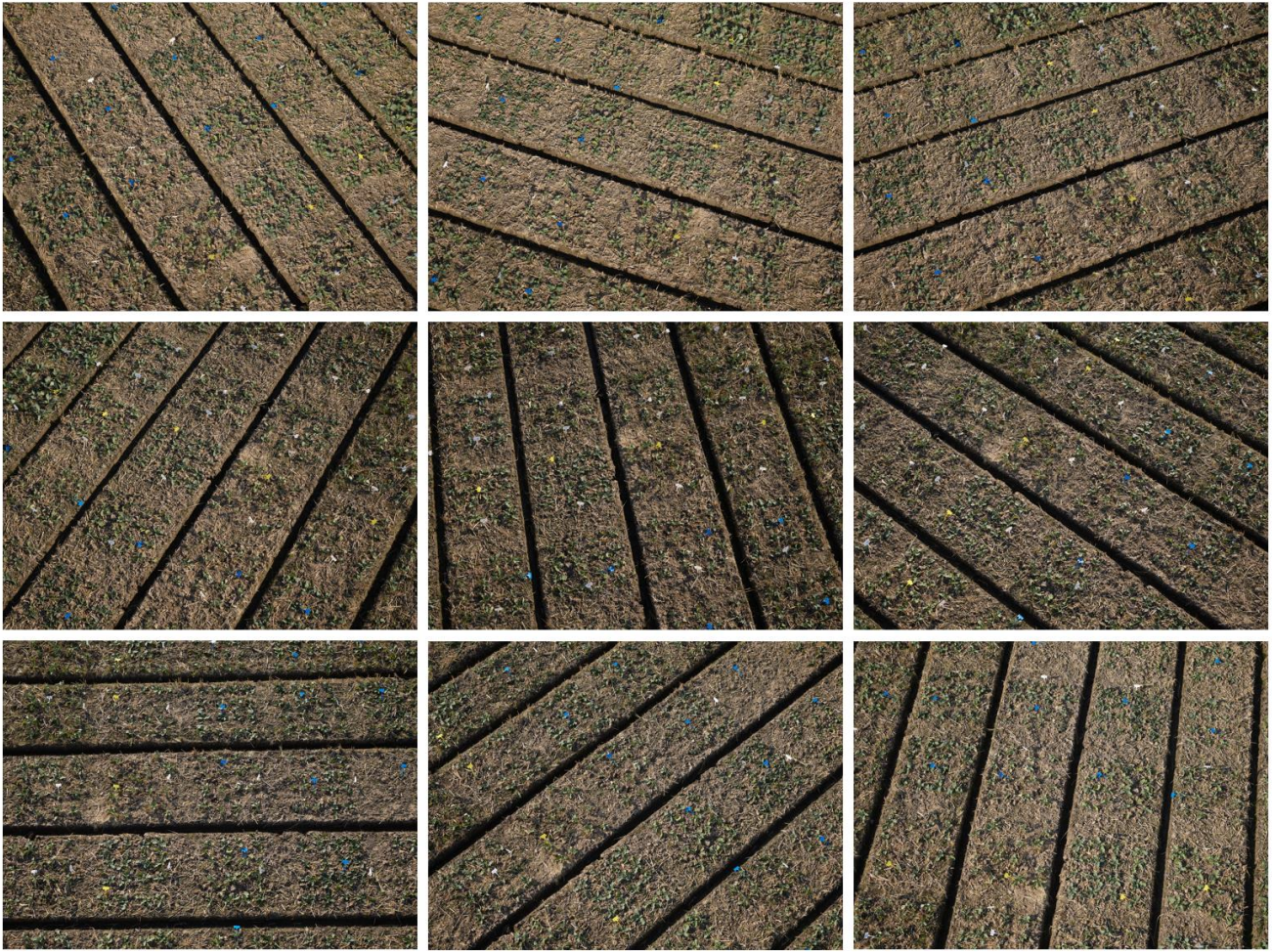

**Fig 3.** Examples of multi-view oblique images from UAV.

**2.3 Field measurements of biomass**

Before the UAV flight plan was initiated, one oilseed rape plant in each of the 300 plots from the second replicate of the 900 plots was randomly selected and marked with a label. Biomass ground truth measurements were conducted immediately after each UAV flight. For each marked plant, the AGB was measured by cutting the plant at the base and weighing the shoot, with the recorded values representing the ground truth biomass for each plot. Biomass data from four plots with insufficient oilseed rape seedlings were excluded from the analysis. Additionally, three plots were removed as the plants had prematurely entered the flowering stage by December 21, 2023. In total, ground truth biomass measurements were obtained for 293 plots.

**2.4 Extraction of plot crop volume and individual crop volume**

2.4.1 Data preprocessing

The IIQ format files were first processed using Capture One (version 12, Phase One A/S, Copenhagen, Denmark), where exposure adjustments were made for any incorrectly exposed images. The corrected images were then exported in JPG format for further processing. Agisoft Metashape (version 1.6.2, Agisoft LLC, St. Petersburg, Russia) was used to process the JPG images, following a sequence of steps: Align Photos, Build Dense Point Clouds, Build Mesh, Build Orthomosaic, Build DEM (for bare soil), and Build digital terrain model (DTM) (for seedling-stage oilseed rape). Once the orthomosaic, DEM, and DTM were generated, they were exported with a fixed geospatial bounding box and ground sampling distance (GSD) to ensure pixel-level correspondence among the three outputs (Wan et al., 2018).

Next, in MATLAB (version 2021a, MathWorks Inc., USA), the DTM of seedling-stage oilseed rape was subtracted from the DEM of bare soil to generate the crop surface model (CSM) for the seedling-stage oilseed rape. The orthomosaic was then automated segmented by Plot-Seg model (Lu et al., 2024). Each plot was cropped into 2400 × 2400 pixels sub-images, and the CSM was similarly cropped into sub-CSMs as shown in the step 1 in Fig 4. The sub-images and sub-CSMs were aligned to ensure pixel-level correspondence for subsequent analysis (Wan et al., 2020).

2.4.2 Instance segmentation

To accurately identify and segment individual oilseed rape plants within each plot, the pre-trained Mask-RCNN instance segmentation model from a previous study was employed (Shen et al., 2024). This model, originally trained using 90 images for training, 30 for validation, and 60 for testing, was specifically designed to distinguish oilseed rape plants. The training dataset had been augmented through various techniques, including color transformations, image flipping, rotation, scaling, and noise injection, enhancing the model's robustness. Color transformations included brightness, contrast, and saturation adjustments, with parameters set between 0.5 and 1.5.

The Mask-RCNN algorithm operates in two stages. In the first stage, potential regions containing targets are proposed, and in the second stage, these regions are refined to produce bounding boxes, masks, and classifications. ResNet50 was used as the backbone for feature extraction, and a feature pyramid network (FPN) was implemented to create feature maps that combined high-resolution and deep semantic information. These feature maps were passed to a region proposal network (RPN), which scans the images for regions of interest (anchors) that might contain oilseed rape plants.

Since region proposals can vary in size, ROI Align was used to standardize their dimensions by performing bilinear interpolation. The resized regions were processed by three branches: one for generating masks, one for identifying the target class (oilseed rape and label in this study), and the third for determining the bounding boxes. The results from these branches were combined to achieve instance segmentation, with each individual oilseed rape plant being accurately identified and assigned a distinct color while the label was marked in white.

2.4.3 Volume extraction and biomass estimation

The extraction workflow for plot crop volume and individual crop volume is shown in Fig 4, after obtaining the plot oilseed rape segmentation result from the Mask-RCNN model, the image was imported into MATLAB and converted into a binary format. In this binary image, pixels corresponding to all of the identified oilseed rape plants were assigned a value of 1, while pixels corresponding to the background or labels were set to 0. The corresponding sub-CSM was also loaded into MATLAB. To calculate the plot crop volume, the binary image was element-wise multiplied with the sub-CSM, and the sum of the resulting pixel values was computed as shown in Equation (1) and Equation (2). This sum represented the cumulative height of all oilseed rape plants within the plot, serving as the plot crop volume for that specific plot.

$$P(x,y) = \begin{cases} 1 & \text{for oilseed rape pixels} \\ 0 & \text{for background or label pixels} \end{cases} \quad (1)$$

$$V_{plot} = \sum_{x,y} P(x,y) \times CSM(x,y) / N \quad (2)$$

where $P(x,y)$ represents the binary segmentation result from the Mask-RCNN model, $CSM(x,y)$ represents the corresponding height values in the sub-CSM, $V_{plot}$ is the total plot crop volume, $N$ is the number of oilseed rape plants in the plot (in this case, 16).

To identify the oilseed rape plant closest to the label, a distance-based algorithm was applied. First, the white label was detected in the image, and the coordinates $T(x_t, y_t)$ of its tip were determined. To find the tip of the label, analyze the contour coordinates. The tip of the label is the point that has the maximum distance from the geometric center of the label. The geometric center can be calculated by Equation (3) and Equation (4) as follows. This point typically represents the sharpest corner of the label and can be identified by calculating the distance of each contour point from the centroid of the label as Equation (5). Next, the centroids of each individual oilseed rape plant instance were calculated using standard centroid detection methods. For each instance, the Euclidean distance between the plant's centroid $C_i(x_{rape,i}, y_{rape,i})$ and the tip of the label was computed as Equation (6). The oilseed rape plant with the minimum distance to the label's tip was selected as the target plant.

$$x_c = \frac{1}{n} \sum_{i=1}^{n} x_{label,i} \quad (3)$$

$$y_c = \frac{1}{n} \sum_{i=1}^{n} y_{label,i} \quad (4)$$

$$d_{label,i} = \sqrt{(x_c - x_{label,i})^2 + (y_c - y_{label,i})^2} \quad (5)$$

$$d_{target,i} = \sqrt{(x_{tip} - x_{rape,i})^2 + (y_{tip} - y_{rape,i})^2} \quad (6)$$

where $C(x_c, y_c)$ represents the geometric center of the label, $d_{label,i}$ represents the Euclidean distance between the geometric center of the label and each contour point $(x_{label,i}, y_{label,i})$, $d_{target,i}$ represents the Euclidean distance between the label tip $(x_{tip}, y_{tip})$ and each plant's centroid $C_i(x_{rape,i}, y_{rape,i})$.

After extracting the target oilseed rape from the segmentation process, the image was imported into MATLAB and converted into a binary format. In this binary image, pixels corresponding to the identified target oilseed rape plant were assigned a value of 1, while all other pixels were set to 0. The corresponding sub-CSM was also loaded into MATLAB. To calculate the individual crop volume, the binary image was element-wise multiplied with the sub-CSM, and the sum of the resulting pixel values was computed, as shown in Equation (7) and Equation (8). This sum represented the cumulative height of all the pixels corresponding to the target oilseed rape plant, serving as the individual crop volume for that specific plant.

$$P_{ind}(x,y) = \begin{cases} 1 & \text{for target oilseed rape pixels} \\ 0 & \text{for all other pixels} \end{cases} \quad (7)$$

$$V_{ind} = \sum_{x,y} P_{ind}(x,y) \times CSM(x,y) \quad (8)$$

where $P_{ind}(x,y)$ represents the binary segmentation result for the target oilseed rape, $CSM(x,y)$ represents the corresponding height values in the sub-CSM, $V_{plot}$ is the total individual crop volume.

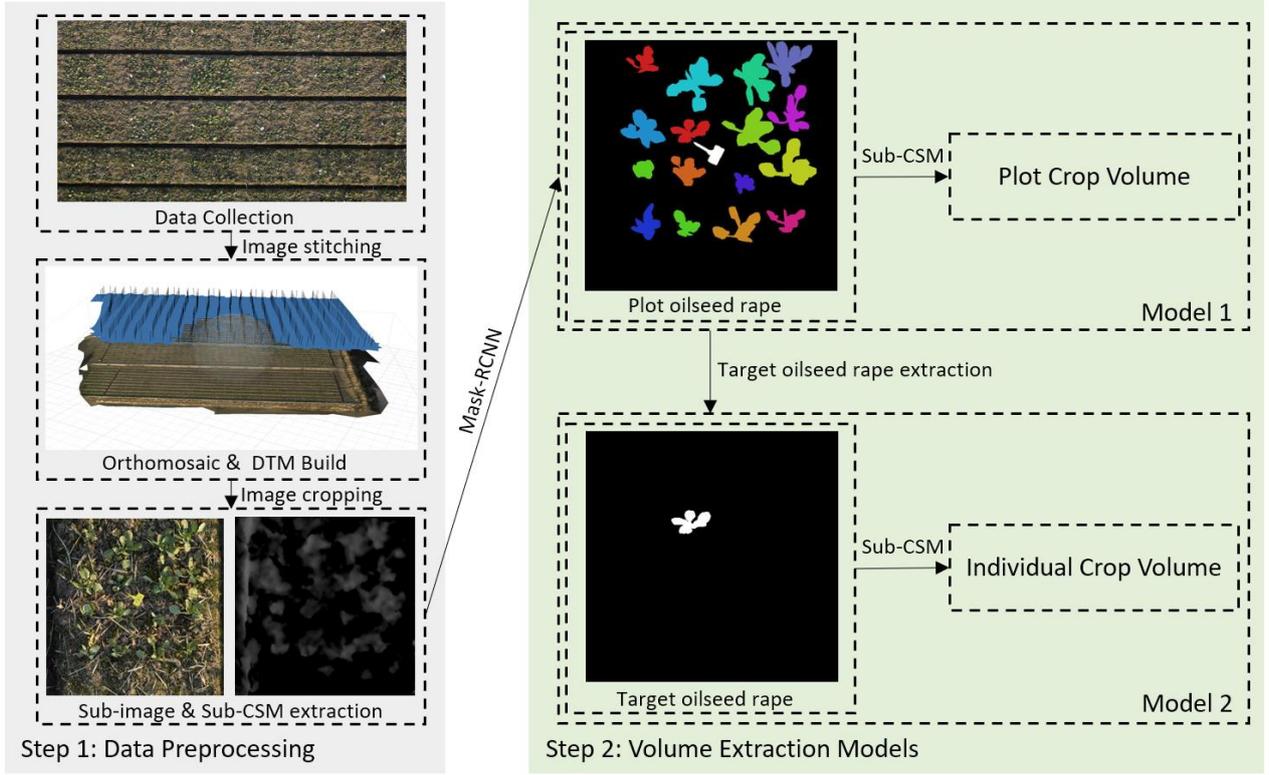

**Fig 4**. The workflow of the extraction model for plot crop volume and individual crop volume based on vertical photography images.

Based on the previously described steps, $V_{plot}$ and $V_{ind}$ were derived to represent the envelope volumes of the entire oilseed rape plot and the individual target oilseed rape plant, respectively. Both $V_{plot}$ and $V_{ind}$ are expressed in cubic centimeters (cm³). To estimate the biomass of the oilseed rape plants, we used a simple linear regression model to fit the relationship between these envelope volumes and the measured biomass of the target oilseed rape plant. Once the biomass estimation model was established, we quantitatively evaluated the accuracy of the estimates using three metrics: $R^2$, RMSE, and mean absolute percentage error (MAPE).

$$R^2 = 1 - \frac{\sum_{i=1}^{n}(B_{true,i} - B_{pred,i})^2}{\sum_{i=1}^{n}(B_{true,i} - \bar{B}_{true})^2} \quad (9)$$

$$RMSE = \sqrt{\frac{1}{n}\sum_{i=1}^{n}(B_{true,i} - B_{pred,i})^2} \quad (10)$$

$$MAPE = \frac{100\%}{n}\sum_{i=1}^{n}\left|\frac{B_{true,i} - B_{pred,i}}{B_{true,i}}\right| \quad (11)$$

where $B_{true,i}$ represents the true measured biomass for the *i*-th plant, $B_{pred,i}$ is the predicted biomass based on the model, $\bar{B}_{true}$ is the mean of the true biomass values, *n* is the total number of samples (in this case, 293).

## 2.5 Point cloud volume extraction

2.5.1 3DGS for 3D reconstruction

To perform 3D reconstruction of oilseed rape plants from 36 multi-view oblique UAV images, the 3DGS method was employed. 3DGS is a lightweight and efficient technique that represents point clouds as Gaussian splats, providing smooth and continuous surface reconstructions without the need for computationally expensive volumetric grids or neural network-based optimizations (Gao & Qi, 2024).

The 3DGS pipeline begins by loading the multi-view oblique RGB images captured from the UAV platform. These images were preprocessed through camera pose estimation and image alignment. The 3DGS algorithm uses the estimated camera poses and depth information extracted from the images to initialize a point cloud representation of the scene. Unlike traditional volumetric-based methods, which require computationally expensive volumetric rendering, 3DGS employs Gaussian splats to represent the point cloud. Each point in the cloud is modeled as a Gaussian distribution with a spatial location, variance, and color attributes. This allows for efficient and accurate rendering of the 3D scene, capturing fine details of the plant structure without the need for high computational costs associated with other methods (Wu et al., 2024).

Through a projection process, each 3D point is represented as a Gaussian splat and projected onto the 2D image plane based on the camera's intrinsic and extrinsic parameters. This projection maintains the 3D positional accuracy of the splats and their relative contributions in the scene. For each pixel in the input image, the algorithm calculates the contribution of each Gaussian splat to the image space using a projection model. These splats are blended together in the image space based on their overlapping regions, producing a smooth and continuous reconstruction of the oilseed rape canopy. The point cloud is further refined by adjusting the size and variance of each Gaussian splat to minimize reconstruction errors and improve surface representation.

To enhance reconstruction quality and computational efficiency, adaptive density control is applied. This mechanism dynamically adjusts the number of Gaussian splats based on the complexity of the scene. In regions requiring finer details—such as dense vegetation—more splats with smaller variances are used, while fewer, larger splats are employed in simpler areas like the ground. Additionally, the method leverages a differentiable tile rasterizer, which handles overlapping splats and ensures smooth transitions between them in the image space. This tile-based approach accelerates rendering by dividing the image into smaller regions for parallel processing, while the differentiability of the rasterizer allows for fine adjustments to splat properties through backpropagation, ensuring an accurate 3D reconstruction (Kerbl et al., 2023).

The quality of the 3D reconstruction generated by 3DGS was validated using Peak Signal-to-Noise Ratio (PSNR). PSNR is a widely-used metric for evaluating the quality of reconstructions, particularly in image-based tasks. It measures the ratio between the maximum possible power of a signal (in this case, the reference or ground truth model) and the power of corrupting noise (in this case, the reconstruction error). A higher PSNR value indicates better reconstruction quality, as it implies a smaller difference between the reconstructed scene and

the reference. The PSNR is calculated as follows:

$$PSNR = 10 * log_{10}\left(\frac{MAX_I^2}{MSE}\right) \quad (12)$$

$$MSE = \frac{1}{N}\sum_{i=1}^{N}\left(I_{true}(i) - I_{reconstructed}(i)\right)^2 \quad (13)$$

where $MAX_I$ is the maximum possible pixel value of the image, which is 255 in this study as we use 8-bit image here. MSE represents the mean squared error, which is the average of the squared differences between corresponding pixels in the reference (ground truth image) and the reconstructed images. $I_{true}(i)$ is the pixel intensity of the ground truth image, $I_{reconstructed}(i)$ is the pixel intensity of the reconstructed image, $N$ is the total number of pixels.

Additionally, to compare the effectiveness of 3D reconstruction, we selected conventional point clouds extracted from Structure-from-Motion (SfM), as well as point clouds generated using instant-NGP (the fastest NeRF algorithm) and Mip-NeRF (the most accurate NeRF algorithm). All data processing was performed on a desktop computer equipped with an Intel Core i9-14900KF CPU, an NVIDIA GeForce RTX 3090 GPU, and 128GB of 2666MHz DDR4 RAM.

2.5.2 Segmentation of 3DGS point cloud integrated with SAM

In this study, we integrated the SAM with 3DGS to develop a precise 3D point cloud segmentation method for oilseed rape plants. Initially, multi-view oblique UAV images were collected and used for camera pose estimation. These images were then fed into the 3DGS pipeline, where the 3D structure of the scene was represented as Gaussian splats. Each point in the point cloud was modeled as a Gaussian with specific attributes such as spatial location, variance, and color.

SAM was employed to generate 2D segmentation masks from the UAV images. These segmentation masks were produced at multiple granularities to capture the varying sizes and shapes of individual plants within the field. The next critical step involved projecting these 2D segmentation results into the 3D space of the Gaussian splats, and then we got the SAM mask. By aligning the 2D features generated by SAM with the 3D point cloud, each Gaussian splat was assigned a corresponding feature vector, enabling the 3D segmentation of oilseed rape plants (Ding et al., 2024).

To improve the consistency of the segmentation across multiple views, contrastive learning was applied. This step refined the feature alignment, ensuring that the segmentation features from SAM maintained coherence when mapped into the 3D space, even from different angles and perspectives. This feature refinement was critical in achieving accurate segmentation in the 3D point cloud.

Following the projection of SAM's 2D segmentation results into the 3D space, additional post-processing techniques were applied to further enhance the accuracy of the segmentation. Region-growing algorithms were used to extend and unify segments, while noise filtering was employed to eliminate any erroneous points or outliers that might have been introduced during the projection or segmentation process. The final segmented 3D point cloud provided a precise representation of individual 3D oilseed rape instance, allowing for high-throughput phenotyping in the field. This integrated approach combined the high-quality 2D segmentation capabilities of SAM with the efficiency and scalability of 3DGS, enabling accurate plant detection and segmentation across large-scale agricultural environments as shown in Fig 5.

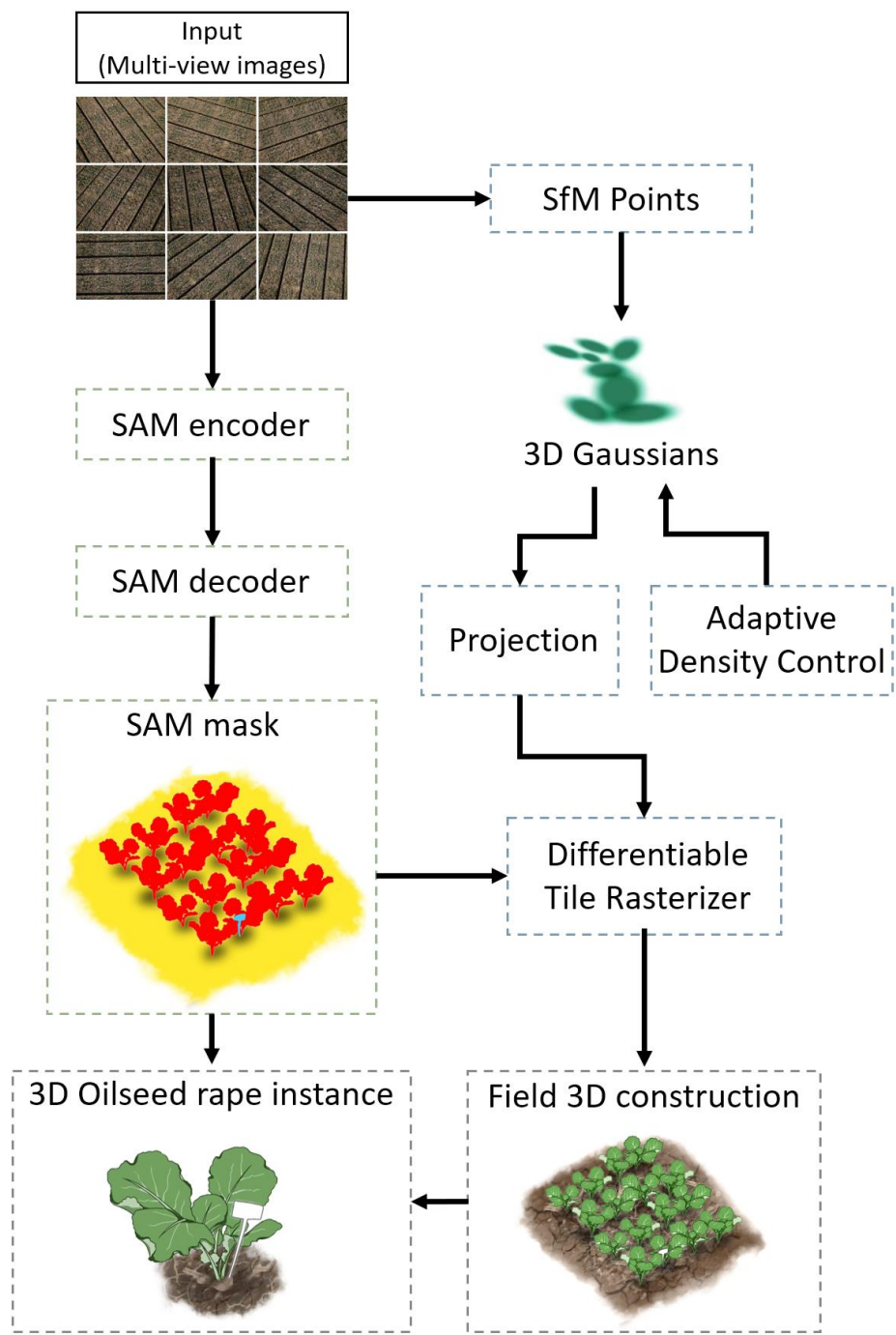

**Fig 5.** The architecture of the 3D oilseed rape instance segmentation based on 3DGS and SAM.

After performing the segmentation of 3DGS point cloud, the segmented point clouds were projected back onto the original 36 UAV images. This projection was done along the same viewing angles as the original images, allowing the segmented 3D point cloud to be mapped accurately onto the 2D planes of each image. To evaluate the effectiveness of the segmentation model, the mean intersection over union (mIoU), precision, recall and F1-score were calculated by comparing the projected segmentation results with the corresponding target oilseed rape regions in the original UAV images as Equation (15-18). A total of 293 target oilseed rape plants were annotated using the Labelme tool. The plots were manually classified into two growth stages: seedling stage (n=247) and bolting stage (n=46). The mIoU, precision, recall and F1-score values were then calculated separately for each group to assess segmentation performance across different growth stages.

The intersection over union (IoU) is a commonly used metric in image segmentation to measure the overlap between the predicted segmentation mask and the ground truth. For each image, IoU is calculated by dividing the area of overlap between the predicted mask and the ground truth mask by the area of their union. Mathematically, IoU is defined as Equation (14):

$$IoU = \frac{|A_{pred} \cap A_{true}|}{|A_{pred} \cup A_{true}|} \tag{14}$$

$$mIoU = \frac{1}{N}\sum_{i=1}^{N}\frac{|A_{pred,i} \cap A_{true,i}|}{|A_{pred,i} \cup A_{true,i}|} \tag{15}$$

$$Precision = \frac{TP}{TP+FP} \tag{16}$$

$$Recall = \frac{TP}{TP+FN} \tag{17}$$

$$F1-score = 2 * \frac{Precision*Recall}{Precision+Recall} \tag{18}$$

where $A_{pred,i}$ is the pixels of the *i*-th image in the predicted segmentation (in this case, the projection of the segmented 3D point cloud onto the image plane), $A_{true,i}$ is the pixels of the *i*-th image in the ground truth segmentation (the true oilseed rape regions in the original UAV image), *N* is the total number of images (in this case, 36), TP represents the true positives, FP represents the false positives and FN represents the false negatives.

By calculating mIoU, the degree of overlap between the projected 3D point cloud and the ground truth segmentation can be quantified, providing a clear measure of the segmentation model's performance. A higher mIoU value indicates better agreement between the predicted segmentation and the actual oilseed rape regions in the images, confirming the accuracy of the 3DGS point cloud segmentation approach. This combination of projection-based validation and mIoU provides a comprehensive evaluation of the segmentation model, ensuring that it not only performs well in 3D space but also maintains accuracy when projected back into 2D images.

2.5.3 Volume extraction and biomass estimation

As shown in Fig 6, after conducting 75 multi-view oblique flight missions, each consisting of 36 images, data processing was performed sequentially for the 75 groups of images. For each group, 3DGS was used to reconstruct a 3D model of the target scene. Following the 3D reconstruction, the SAM module was employed to extract the SAM masks, which were integrated into the 3DGS process. This integration allowed for an additional point cloud category dimension for each Gaussian point, thus enhancing the reconstruction by associating each Gaussian point with a specific class.

Next, in each image set, the target oilseed rape plants from four experimental plots were identified and segmented using the enhanced 3DGS with the SAM mask. The corresponding Gaussian point clouds were then voxelized to calculate the point cloud volume of the marked oilseed rape plants. To estimate the biomass from the extracted point cloud volumes, a simple linear regression model was applied, establishing a relationship between the point cloud volume and the actual biomass. For evaluating the accuracy of the biomass estimation, the $R^2$, RMSE, and MAPE metrics were calculated, following the same procedures and formulas as outlined in Equations (9-11).

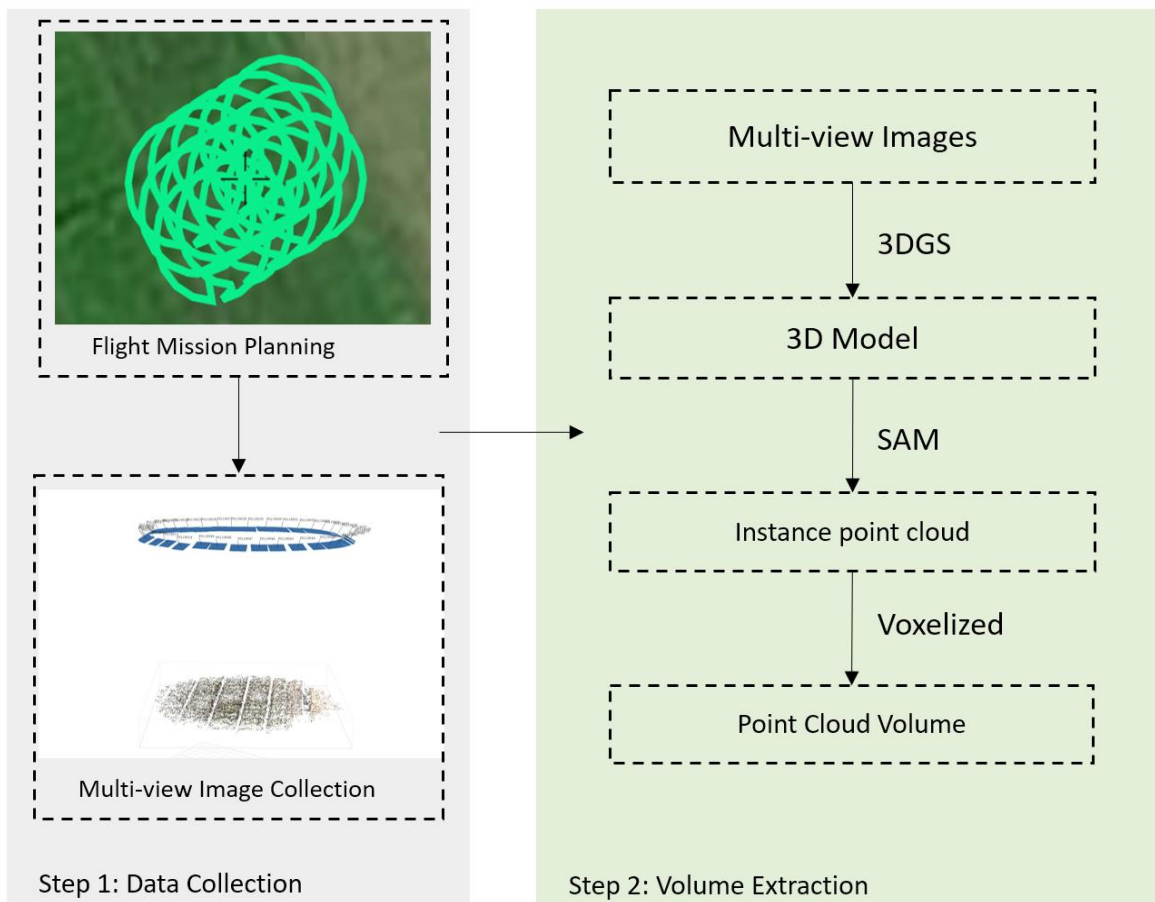

**Fig 6.** The workflow of extraction model of point cloud volume based on multi-view oblique imaging, 3DGS and SAM model.

## 3. Results
### 3.1 Performance of 3D reconstruction

Fig 7(a) illustrates the 3D field reconstruction model generated using 36 multi-view oblique images captured during a single UAV flight, processed by 3DGS. The detailed reconstruction result of an individual plot is shown in Fig 7(b). As depicted in the close-up of the plot, the 3DGS method successfully reconstructs the oilseed rape plants, allowing for the clear identification of leaf count and plant structure. The reconstructed model captures detailed morphological features of the plants and enables high-resolution visualization of plant architecture.

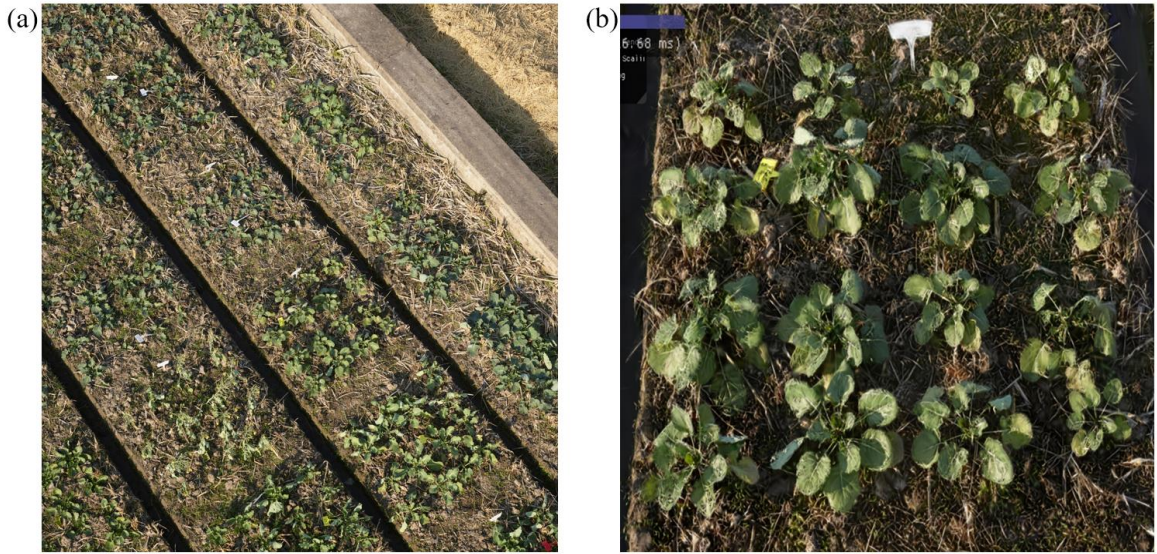

**Fig 7.** 3DGS 3D reconstruction results. (a) Field reconstruction result (b) Individual plot reconstruction result.

In Table 1, the performance of different 3D reconstruction methods is compared in terms of PSNR, training time, and frames per second (FPS). The methods evaluated include SfM, Instant-NGP, Mip-NeRF360, and two configurations of 3DGS (7k and 30k interations). During the first 7k iterations of the 3DGS algorithm, the accuracy of the 3D reconstruction model converges rapidly, achieving good reconstruction quality. After 30k iterations, the model's accuracy converges to its optimal performance. The PSNR values indicate the reconstruction quality, with higher values corresponding to better quality. SfM achieved a PSNR of 21.27, the lowest among the methods, while Mip-NeRF360 demonstrated the highest PSNR of 29.72. Instant-NGP and the two 3DGS configurations also performed well, with PSNR values of 25.18 for Instant-NGP, 27.43 for 3DGS-7k, and 29.53 for 3DGS-30k. In terms of training time, Instant-NGP and 3DGS-7k were the fastest methods, both requiring only 7 minutes for training. 3DGS-30k took longer, requiring 49 minutes, while Mip-NeRF360 had the longest training time at 45 hours. SfM required a moderate training time of 12 minutes. The 3D reconstruction accuracy of 3DGS-30k is very close to that of Mip-NeRF360, while significantly outperforming the traditional SfM method. Moreover, 3DGS-30k offers a substantial advantage in reconstruction speed over Mip-NeRF360. Among the fast reconstruction methods, 3DGS-7k and Instant-NGP, 3DGS-7k not only completes the process more quickly but also delivers superior reconstruction quality compared to Instant-NGP. FPS is a crucial metric for evaluating the real-time rendering speed of various algorithms in 3D reconstruction visualization. In this case, SfM does not have an FPS value because it generates a point cloud without real-time visualization. The 3DGS method stands out from all others due to its use of Gaussian splatting, which eliminates the need for volumetric rendering, resulting in a significant performance advantage. 3DGS-7k achieved the highest FPS at 139, followed by 3DGS-30k with 91 FPS. Instant NGP, while not as fast as 3DGS, still offers a reasonable FPS of 8.9. Mip-NeRF360, despite its high accuracy, is by far the slowest, producing only 0.074 FPS, making it unsuitable for real-time applications.

**Table 1.** PSNR, training time and FPS for SfM, Instant-NGR, Mip-NeRF360, 3DGS-7k and 3DGS-30k 3D reconstruction

| Methods | PSNR | Training time | FPS |
| --- | --- | --- | --- |
| SfM | 21.27 | 12 min | / |
| Instant-NGP | 25.18 | 7 min | 8.9 |
| Mip-NeRF360 | 29.72 | 45 h | 0.074 |
| 3DGS-7k | 27.43 | 7 min | 139 |
| 3DGS-30k | 29.53 | 49 min | 91 |

**3.2 Performance of 3DGS point cloud segmentation**

Fig 8(a) presents the 3D Gaussian point cloud of the field reconstructed using 36 images obtained from a multi-view oblique UAV flight. This large-scale 3D point cloud captures detailed spatial information about the entire field. After applying the point cloud segmentation process, four target oilseed rape plants, which were marked with labels during the experiment, were successfully extracted, as shown in Fig 8(b). Fig 8(c) shows the 3D Gaussian point cloud segmentation result for one of the target oilseed rape plants. After segmentation, the 3D point cloud was rendered into 2D RGB images from 36 different angles, as illustrated in Fig 8(d). The detailed projections allow for a comprehensive visualization of the plant structure, demonstrating the ability of the 3DGS algorithm to capture the fine details of the individual oilseed rape plant from multiple perspectives.

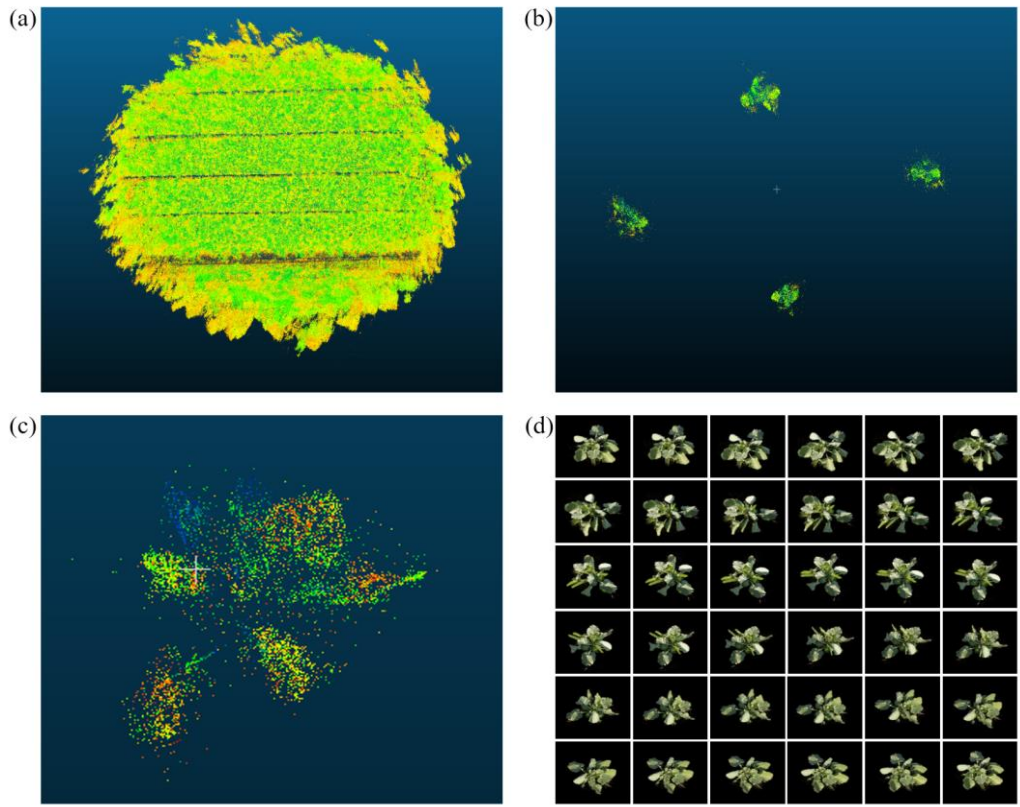

**Fig 8.** 3D Gaussian point cloud and 3DGS point cloud segmentation results. (a) Field 3DGS point

cloud (b)Target oilseed rape point cloud. (c) Segmented point cloud of individual oilseed rape (d) Rendered projections of segmented point cloud from 36 angles.

The performance of 3DGS point cloud segmentation was quantitatively evaluated using the metrics of mIoU, Precision, Recall, and F1-score, as presented in Table 2. These metrics were computed by comparing the rendered projections of the segmented point cloud with manually labeled target oilseed rape plants across two distinct growth stages: seedling (n=247) and bolting (n=46). The segmentation results for the seedling stage showed an mIoU of 0.966, with a Precision of 98.14%, Recall of 98.46%, and an F1-score of 0.983. For the bolting stage, the mIoU slightly decreased to 0.931, with a Precision of 96.94%, Recall of 95.96%, and an F1-score of 0.964. When considering both growth stages together, the overall mIoU was 0.961 and the combined Precision, Recall, and F1-score for all 293 plants were 97.95%, 98.05%, and 0.980, respectively.

**Table 2.** mIoU, Precision, Recall, and F1-score for the comparison between rendered projections of segmented point cloud and manually labeled target oilseed rape. The evaluation set consists of 293 plants, including 247 at the seedling stage and 43 at the bolting stage.

| Growth stage | mIoU | Precision | Recall | F1-score |
| --- | --- | --- | --- | --- |
| Seedling stage | 0.966 | 98.1% | 98.5% | 0.983 |
| Bolting stage | 0.931 | 96.9% | 96.0% | 0.964 |
| Whole 2 stages | 0.961 | 98.0% | 98.1% | 0.980 |

### 3.3 Performance of three biomass estimation models

The results of the biomass estimation using three different volume models—plot volume, individual crop volume, and point cloud volume—are presented in the scatter plots shown in the figure. The first plot illustrates the relationship between the plot volume model and the measured biomass. As shown in Fig 9(a), the correlation was weak, with an $R^2$ value of 0.009, an RMSE of 18.14 g/plant, and a MAPE of 61.65%. This suggests that the plot volume model alone is not a reliable predictor of biomass. In contrast, Fig 9(b) shows a much stronger correlation between the individual crop volume model and biomass, yielding an $R^2$ value of 0.790, an RMSE of 8.35 g/plant, and a MAPE of 23.76%. This demonstrates that the individual crop volume model provides a significantly better estimation of biomass compared to the plot volume model. Fig 9(c) displays the results for the point cloud volume model, which achieved the highest accuracy in biomass estimation. The $R^2$ value reached 0.976, with an RMSE of 2.82 g/plant and a MAPE of 6.81%, indicating a strong fit between the predicted and actual biomass. These results clearly show that the point cloud volume model, which incorporates detailed 3D structure information, offers the most accurate biomass estimation among the three models.

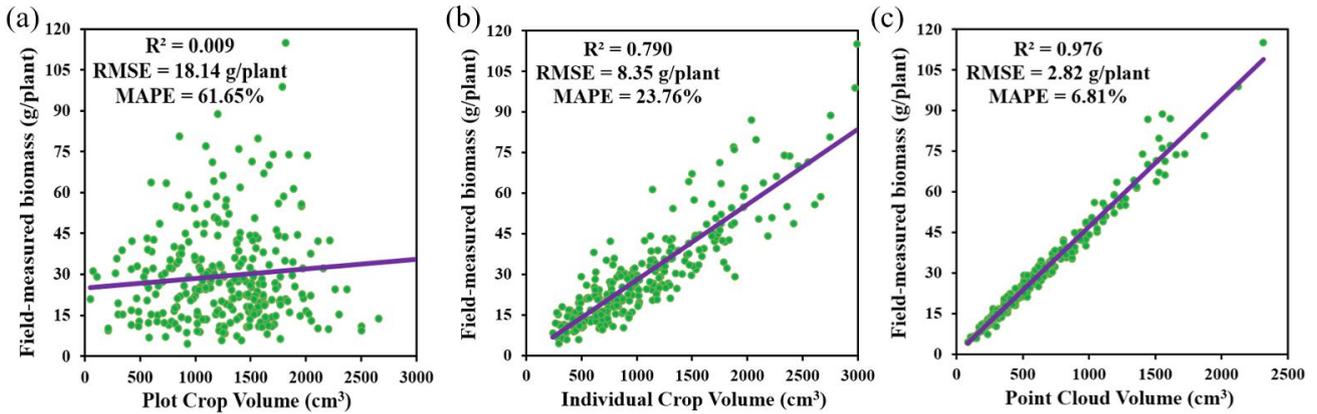

**Fig 9.** Correlation between oilseed rape biomass and volume extracted by three models (a) Plot volume model (b) Individual crop volume model (c) Point cloud volume model. The $R^2$, RMSE and rRMSE represent the coefficient of determination, root mean square error and relative RMSE, respectively.

**3.4 Error distribution of point cloud volume model**

Fig 10 presents the residual analysis of the biomass extracted using the point cloud volume model in comparison to the field-measured biomass. The distribution of differences between the UAV-based biomass estimation and the field-measured biomass is shown in Fig 10(a). The residuals tend to cluster around zero, with only a few outliers, indicating that the point cloud volume model generally performs well in estimating biomass. However, there are some notable deviations for field-measured biomass values above 40 g. Fig 10(b) provides a histogram displaying the frequency distribution of the errors between UAV-based biomass estimates and field-measured values. The distribution approximates a normal distribution, centered around zero, which suggests that the model does not introduce significant systematic bias. Most of the errors fall within the -5 g to 5 g range, further confirming that the point cloud volume model delivers a reliable estimation of biomass with minimal deviations.

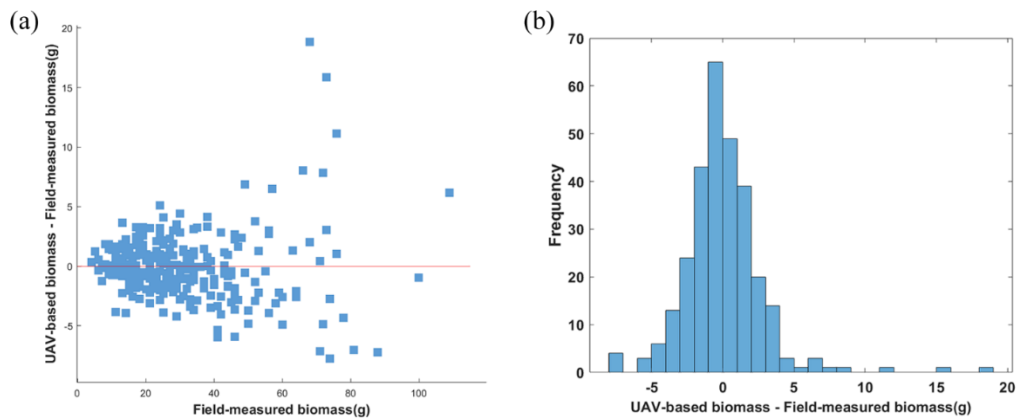

**Fig 10.** Residual analysis of biomass extracted based on point cloud volume model and field-measured biomass. (a)Distribution of differences between UAV-based biomass and field-measured biomass (b)distribution histogram of difference between UAV-based biomass and field-measured biomass.

## 4. Discussion
### 4.1 Comparison of SfM, NeRF and 3DGS

Although the 3DGS reconstruction method offers a more faster and more accurate performance compared to SfM and NeRF, these methods still offer unique strengths and weaknesses, and their performance can be analyzed based on the context of 3D plant reconstruction. From a fundamental standpoint, SfM relies on detecting key points across multiple images, using these to construct a sparse 3D point cloud and eventually forming a dense reconstruction. This method, while widely used, often suffers from limitations in complex scenes where texture is lacking, leading to incomplete reconstructions. On the other hand, NeRF, particularly its more advanced variants like Mip-NeRF360 and Instant-NGP, utilize neural networks to represent the scene in a volumetric manner (Barron et al., 2021). NeRF can produce highly detailed reconstructions, especially in high-complexity environments like vegetation, though it comes with a significant computational cost. 3DGS, as an innovative alternative, offers a more efficient approach. By modeling each point in the scene as a Gaussian splat, this method provides smooth and continuous surfaces while avoiding the heavy computational load associated with traditional volumetric methods (Mazurowski et al., 2023).

Although SfM demonstrated the lowest PSNR in our evaluation, it remains a highly accessible method due to its simplicity and the fact that it does not require specialized hardware or extensive computational resources. In scenarios where large-scale reconstructions are needed and fine details are less critical, such as image stitching for canopy vegetation index extraction, SfM's ability to produce sufficiently accurate results with relatively low computational cost makes it a practical choice. Furthermore, SfM can be effectively applied in low-complexity environments where fine details like individual leaves or small plant structures are not the focus. Additionally, SfM algorithms are already integrated into commercial software like Agisoft Metashape, Smart 3D, and DJI Terra, enabling users to easily and quickly get started with lower learning curves compared to NeRF and 3DGS.

On the other hand, NeRF methods, including Instant-NGP and Mip-NeRF360, excel in capturing intricate details and producing highly photorealistic models (J. Lin, 2024). Despite their higher computational cost, NeRF-based methods outperform 3DGS in scenarios requiring complex occlusions, and view-dependent effects, such as glossiness and reflectivity. For example, in environments where lighting conditions vary significantly across images or where the surfaces have complex, non-Lambertian reflectance properties (like glossy or shiny surfaces), NeRF's volumetric rendering offers superior accuracy. In contrast, 3DGS may face challenges in capturing view-dependent effects or accurately modeling complex light interactions, where NeRF's volumetric methods shine. If photorealistic visualization or view synthesis from arbitrary perspectives is required, NeRF's capacity to model how light behaves in a scene gives it a distinct edge over 3DGS.

To summarize, while 3DGS provides a highly efficient solution for reconstructing large fields and generating accurate, high-speed reconstructions for applications like agriculture and phenotyping, SfM still holds value in lower-detail, large-scale projects due to its simplicity and lower computational demands. Meanwhile, NeRF, especially Mip-NeRF360, outperforms 3DGS in scenarios with highly complex lighting conditions or severe vegetation occlusions. In our 3D reconstruction of seedling-stage oilseed rape, a high level of modeling precision and fast processing speed were required, while complex lighting conditions and vegetation occlusion were not significant concerns. As a result, 3DGS proved to be the optimal choice for this task.

### 4.2 Comparison of three biomass extraction models

In our study, the plot crop volume model performed poorly, especially for oilseed rape, a crop known for its high within-plot heterogeneity. This model is more suited for crops like rice and wheat, where the canopy structure is more uniform. For these crops, the plot crop volume model can provide accurate biomass estimates with relatively low computational cost, making it practical for large-scale applications where fine detail is less critical. On the other hand, the individual crop volume model showed a significant improvement in accuracy. By focusing on the volume of individual oilseed rape plants, it achieved a closer correlation with actual biomass because the estimated volume corresponded directly to the plant being weighed. Moreover, this method does not require time-consuming 3D reconstruction, allowing for rapid biomass extraction—about 60 minutes to process 300 plots. This makes the individual crop volume model a strong option when a high level of precision is not necessary, but speed and efficiency are prioritized.

Lastly, the point cloud volume model provided the highest accuracy, offering a detailed and precise estimation of biomass. However, this method is much slower due to the need for full 3D modeling of each plot. On average, it takes around two minutes per plot, which becomes a bottleneck when working with larger datasets. Although it excels in precision, the point cloud volume model's longer processing time may limit its practical use in high-throughput studies unless the highest accuracy is crucial. Each model has its optimal use case, with the plot crop volume model suiting uniform crops, the individual crop volume model offering a balance of speed and accuracy, and the point cloud volume model providing the highest precision for studies requiring detailed plant structure analysis.

### 4.3 The impact of wind on 3D reconstruction quality

In our experiments, it was observed that the accuracy of point cloud segmentation for oilseed rape plants in the seedling stage was slightly higher than that in the bolting stage. This difference can be attributed to the impact of wind-induced plant movement on the accuracy of 3D reconstruction. When UAVs capture images from 36 different angles, these images are not obtained simultaneously. As a result, if wind causes the plants to sway, their position may shift between different views, leading to discrepancies in 3D reconstruction accuracy. Besides, both the growth stage of the plants and the wind speed at the time of image capture affect the degree of positional displacement caused by wind, ultimately influencing the quality of 3D reconstruction (Sabour et al., 2024).

As illustrated in Fig 11, the PSNR values of the 33 experiments demonstrate a general trend of initial decline followed by recovery. This can be explained by the height of the oilseed rape plants during different growth stages. During the seedling and bolting stages, the plants are shorter, making them less susceptible to wind-induced displacement. However, as the plants enter the flowering stage, they become more prone to wind movement, leading to a drop in 3D reconstruction accuracy. In the later mature stages, the dense growth of the plants and the mutual support between the siliques reduce the impact of wind-induced displacement.

Moreover, wind speed has a more direct influence on the accuracy of 3D reconstruction. In the 33 flights analyzed, wind speeds of levels 2, 3, and 4 were recorded on 9, 14, and 10 days, respectively. The average PSNR values for these wind speeds were 27.03, 25.59, and 24.75, showing a clear correlation between higher wind speeds and lower PSNR values. These results highlight the significant role that wind plays in affecting the precision of 3D reconstruction, with stronger winds leading to greater deviations in the reconstructed model.

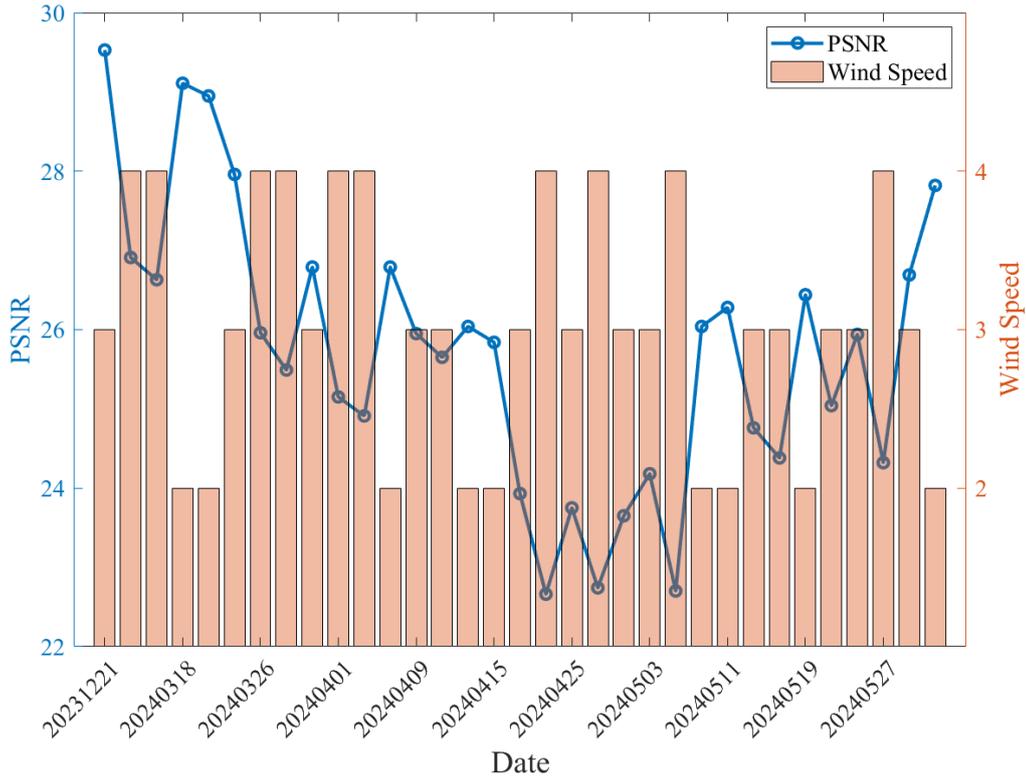

**Fig 11.** The average PSNR of 75 UAV flights conducted each day at varying wind speeds for the 3D reconstruction of oilseed rape using the 3DGS method. The blue line represents the daily average PSNR, while the orange bars indicate the corresponding wind speeds.

**4.4 Implications for future work**

There still exist some limitations in our study, and future research can improve upon these in several directions. One of the primary challenges encountered in this research is the influence of wind on the accuracy of 3D reconstruction, particularly during the flowering stage of oilseed rape or under high wind speeds. Wind can cause disturbances to plant structures, leading to inconsistencies in the captured multi-view oblique images, which are not taken simultaneously, thus affecting the precision of 3D models. To address this issue, future research can explore the development of a 4D Gaussian splatting method that accounts for object movement (Y. Lin et al., 2023), or the creation of new reconstruction algorithms designed to handle blurred images resulting from wind disturbance (Bae et al., 2024). Another promising direction involves the use of coordinated UAV flights, where multiple drones capture multi-view oblique images simultaneously, ensuring that the image sets are taken at the same moment, mitigating the effects of wind on 3D modeling (J. Zhang et al., 2020).

Additionally, while 3DGS has demonstrated significant speed advantages over methods like Mip-NeRF360, the reconstruction process remains a bottleneck when applied to large-scale remote sensing tasks. Further research should focus on optimizing the 3DGS algorithm to enhance its efficiency and reduce processing time for larger datasets, which is crucial for large-scale phenotypic extraction in agricultural applications (Höllein et al., 2024).

Lastly, this study primarily focused on the relationship between crop volume and biomass for oilseed rape, without accounting for the potential differences in plant density across various cultivars. This limitation could affect the accuracy of biomass predictions. Future work could incorporate multispectral imaging or other data sources to refine the model and account for plant density variations, further improving the precision of biomass extraction models.

## 5. Conclusions

This study utilized UAV multi-view oblique imaging combined with advanced 3D reconstruction techniques, including 3DGS, NeRF, and SfM, to assess the accuracy of biomass estimation in oilseed rape. Among the three methods, 3DGS demonstrated a superior balance of speed and precision, making it the most suitable approach for high-throughput phenotyping applications. 3DGS provided high-quality 3D reconstructions with a PSNR value of 29.53 after 30,000 iterations, closely matching the results achieved by Mip-NeRF360 (PSNR 29.72) while significantly outperforming SfM, which yielded a PSNR of 21.27. Moreover, 3DGS offered faster reconstruction speeds than NeRF, with 3DGS-7k achieving 7 min and 3DGS-30k reaching 49 min, making it a viable option for real-time applications where fast processing is required. In terms of biomass estimation, we compared three extraction models: plot crop volume, individual crop volume, and point cloud volume. The point cloud volume model, utilizing 3DGS, provided the highest accuracy with an $R^2$ of 0.976 and a MAPE of 6.81%, though it required longer data processing times, making it ideal for applications where precision is prioritized over speed. The study also identified the impact of wind on 3D reconstruction quality, with wind-induced plant movement reducing PSNR, particularly during the flowering stage or at higher wind speeds. Future work should focus on mitigating these effects by developing 4D Gaussian splatting techniques or employing UAV swarm imaging to capture synchronized multi-view oblique datasets, reducing the impact of wind on plant displacement. Additionally, further optimization of 3DGS is necessary to enhance its processing speed for larger-scale applications. Lastly, future models should incorporate multispectral data to account for crop density variations across different oilseed rape varieties, further improving biomass estimation accuracy.

## Author contribution

Author contributions: **Yutao Shen** contributed to the experimental design, model construction, data collection and manuscript writing. **Hongyu Zhou** and **Xin Yang** contributed to the experimental design, model construction and data collection. **Ziyue Guo** and **Xuqi Lu** contributed to the experimental design and data collection. **Lixi Jiang** provided the experimental materials and facilities. **Yong He** provided suggestions on data processing and manuscript writing. **Haiyan Cen** supervised the experiments at all stages and revised the manuscript. All authors read and approved the final manuscript.

**Declaration of competing interest**

The authors declare that they have no competing financial interests or personal relationships that could have appeared to influence the work reported in this paper.

**Data availability**

Data will be made available on request.

**Acknowledgement**

This work was funded by the National Key R & D Program of China (2021YFD2000104) and the Key R & D Program of Zhejiang Province, China (2021C02057). We extend our heartfelt gratitude to Ruisen Wang, Yi Feng, Zhihong Ma, Pengyao Xie, Yiwen Shao and Mengqi Lyu for their participation in the experiments, and to the Jiaxing Academy of Agricultural Sciences for their invaluable assistance with the experimental data acquisition in 2023-2024.